\documentclass[lettersize,journal]{IEEEtran}
\usepackage{optidef}
\usepackage{amsmath,amsfonts}
\usepackage{algorithmic}
\usepackage{algorithm}
\usepackage{array}
\usepackage{textcomp}
\usepackage{stfloats}
\usepackage{url}
\usepackage{verbatim}
\usepackage{graphicx}
\usepackage{cite}
\usepackage{tabularx}
\usepackage{multirow}
\usepackage{hhline}
\usepackage{subfig}

\begin{document}
\bstctlcite{IEEEexample:BSTcontrol}
\title{Onboard Out-of-Calibration Detection of Deep Learning Models using Conformal Prediction}

\author{Protim Bhattacharjee and Peter Jung 

\thanks{P. Bhattacharjee is with the Department of Optical Sensor Systems, German Aerospace Center (DLR), Berlin, Germany (e-mail: protim.bhattacharjee@dlr.de).}
\thanks{P. Jung is with the Department of Optical Sensor Systems, German Aerospace Center (DLR), Berlin, Germany (e-mail: peter.jung@dlr.de).}%
}

\markboth{Journal of \LaTeX\ Class Files,~Vol.~14, No.~8, August~2021}%
{Shell \MakeLowercase{\textit{et al.}}: A Sample Article Using IEEEtran.cls for IEEE Journals}


\maketitle

\begin{abstract}
The black box nature of deep learning models complicate their usage in critical applications such as remote sensing. Conformal prediction is a method to ensure trust in such scenarios. Subject to data exchangeability, conformal prediction provides finite sample coverage guarantees in the form of a prediction set that is guaranteed to contain the true class within a user defined error rate. In this letter we show that conformal prediction algorithms are related to the uncertainty of the deep learning model and that this relation can be used to detect if the deep learning model is out-of-calibration. Popular classification models like Resnet50, Densenet161, InceptionV3, and MobileNetV2 are applied on remote sensing datasets such as the EuroSAT to demonstrate how under noisy scenarios the model outputs become untrustworthy. Furthermore an out-of-calibration detection procedure relating the model uncertainty and the average size of the conformal prediction set is presented. 
\end{abstract}


\section{Introduction}
\label{sec:intro}
\IEEEPARstart{I}{n} the past decade deep learning (DL) has revolutionized remote sensing with its unprecedented performance on various tasks such as land use and land classification (LULC). It has enabled a multitude of new applications and improved technologies for Earth Observation (EO)~\cite{dl_ai_review} and reduced the time to action for remote sensing downstream tasks. However, deep learning models are still black box in nature and their outputs are not always interpretable, which poses difficulties if these models are to be used in critical infrastructures, safety applications, or are used to derive public and economic policies~\cite{jakob_uncertainty_review}. For example, in case of forest fire detection, it is essential the model provides outputs that the rescue  teams can trust and actions can be taken for efficient use of resources to manage the disaster. 

To ensure performance in accordance with design specifications during deployment, sensor systems, such as cameras and spectrometers, are regularly calibrated~\cite{optSat_DC_book}.
As deep learning models become more prevalent in optical sensor systems, it is natural to also calibrate these models to ensure their performance. Then the following two fold question arises: how do we trust/calibrate the output of DL models and how to identify, in an onboard manner, if the performance of the models is deteriorating. 
\begin{figure*}[!t]
\centering
\subfloat{\includegraphics[width=0.3\textwidth]{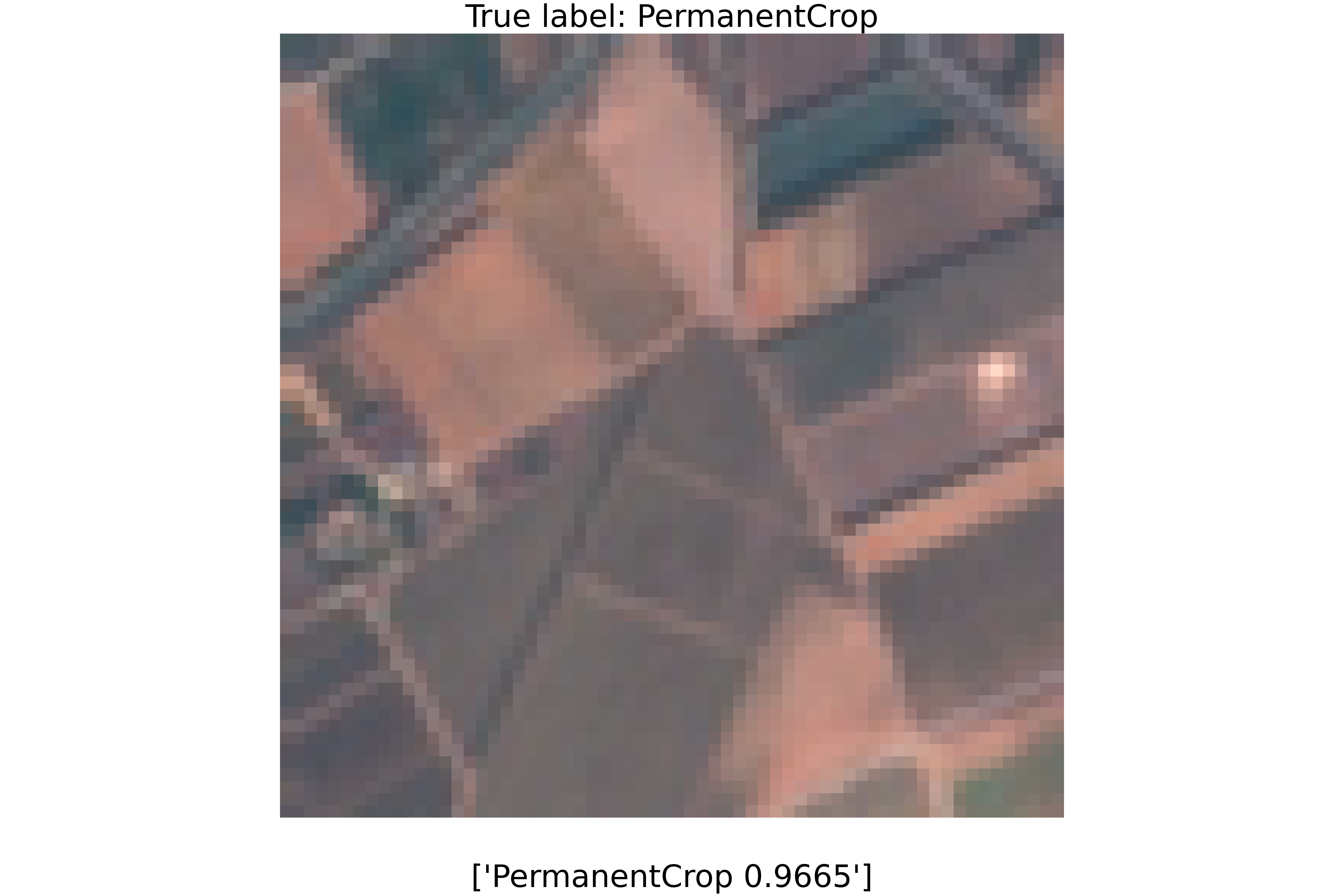}%
}
\hfil
\subfloat{\includegraphics[width=0.3\textwidth]{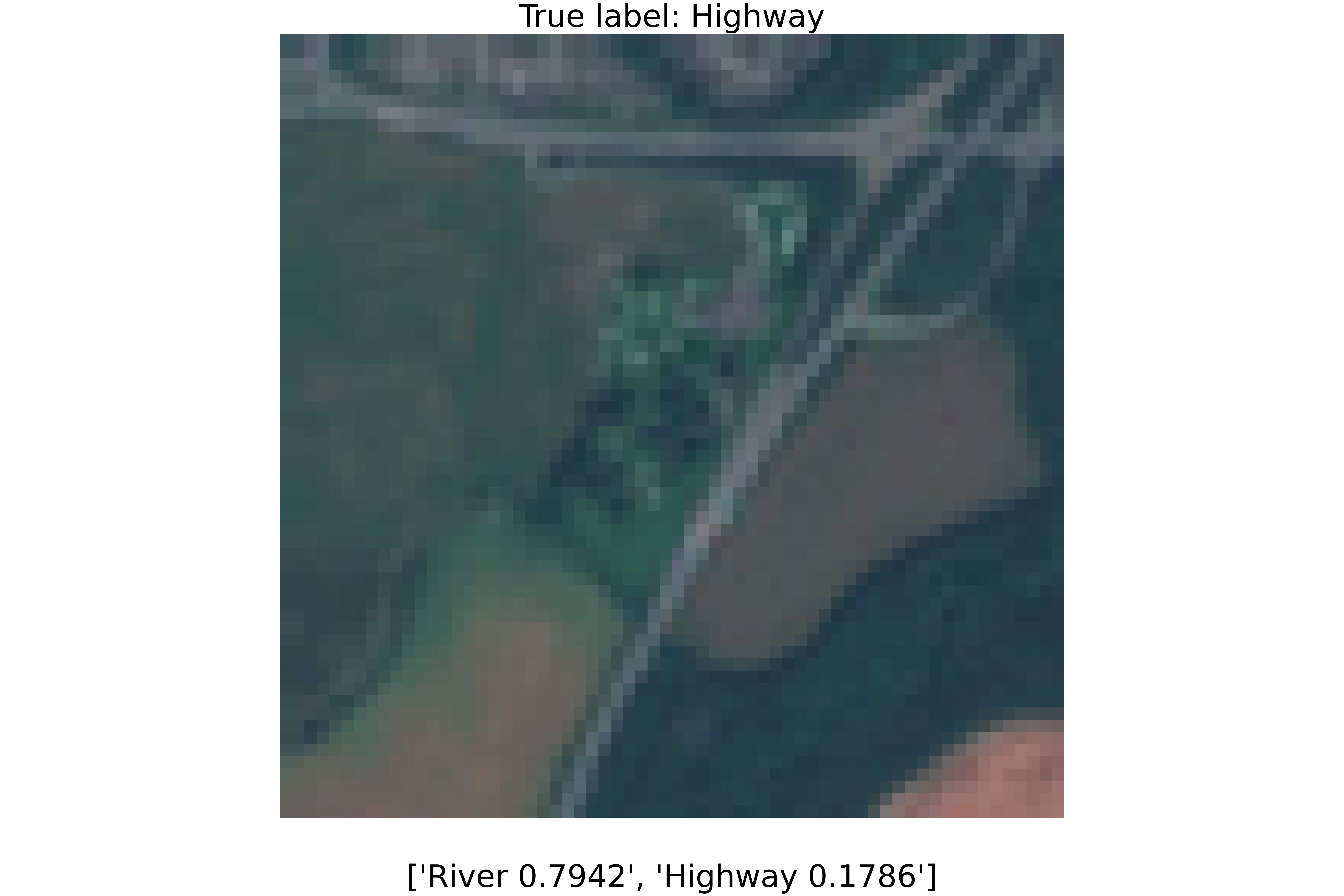}%
}
\hfil
\subfloat{\includegraphics[width=0.3\textwidth]{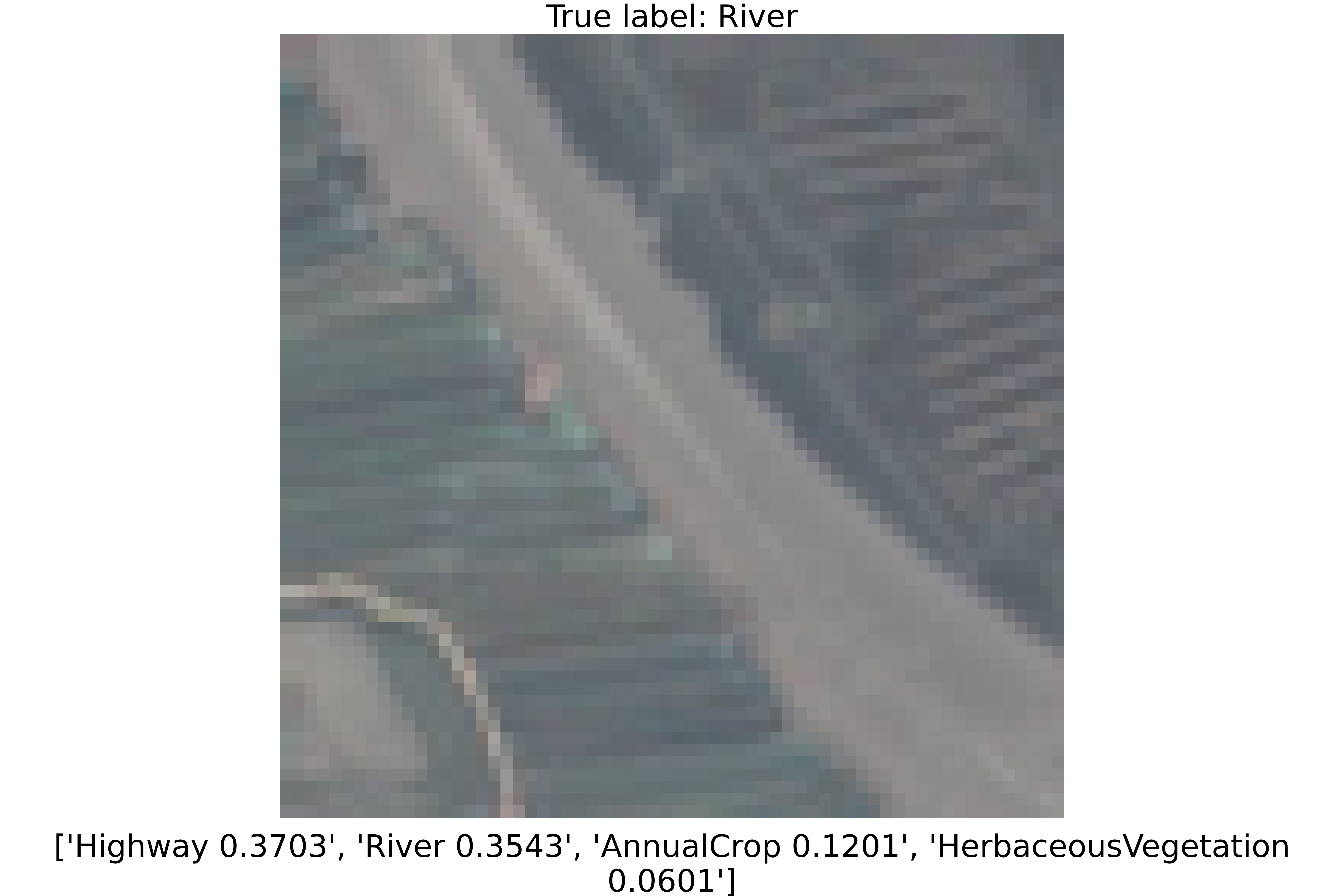}%
}
\caption{Example of conformal prediction using APS~\cite{aps} method with $\epsilon = 0.1$ and Resnet50 for EuroSAT~\cite{eurosat2}. For each image the prediction set along with the softmax output probability is provided. The sum of the softmax values in the prediction sets exceed $1 - \epsilon$. The true class is present in all the prediction sets. For the middle and right panel the point prediction would be incorrect, instead conformal prediction generates a set that contains the true class with probability $1 - \epsilon$.}
\label{cp_example}
\end{figure*}

Trust is established in deep learning models through uncertainty estimation~\cite{jakob_uncertainty_review}. Most of these methods tend to provide uncertainty or calibration estimates for the model during validation but do not provide uncertainty statistics during inference.
They do not provide any indication of whether the model output contains the true class or identify performance deterioration of the model. Conformal Prediction (CP)~\cite{vovk_book} is a method to quantify such trust in deep learning models~\cite{angelopoulos_cp_intro}. Instead of point predictions, that is common for most deep learning models, CP provides, subject to assumptions on the data, a prediction set that is guaranteed to contain the true class within a user defined error rate. It is implemented as a low complexity wrapper around the model and uses a split of the training data, the calibration data, to perform model calibration. The calibration provides a threshold that is representative of the training data and is used to construct prediction sets. Thus, during calibration one can establish an average prediction set size that is representative of the performance of the training data as well as the performance of the model. During inference if the conformalized model outputs larger prediction sets than the calibrated size then one can infer that the model is out-of-calibration. This process is computationally inexpensive and can be used for onboard detection of model performance deterioration. 

CP has been previously used in remote sensing for anomaly detection in  space surveillance~\cite{cp_space_surveillance} and streaming data~\cite{cp_stream_data}; for estimating uncertainty in LULC~\cite{cp_lulc} and in radar target detection recognition~\cite{cp_letters}. 
In this work, CP will be used to detect out-of-calibration models. By studying the relation between model uncertainty and CP we conclude that conformally calibrated uncertain models are easily identified to be out-of-calibration in noisy scenarios and overconfident models cannot be conformalized for such analysis. Thus, the presented work 
\begin{enumerate}
    \item Explores the relationship between conformal prediction and model uncertainty and
    \item Exploits this relationship to perform onboard out-of-calibration detection for deep learning models.
\end{enumerate}

\section{Conformal prediction}
\label{sec:cp}

Conformal prediction~\cite{angelopoulos_cp_intro} is a finite sample coverage guarantee for recognition of the correct class for a given model. It provides a \textit{prediction set} $\mathcal{C}$, instead of a single (point-) prediction, of possible classes to which the test data may belong subject to user/application defined error rate, $\epsilon$. For a test data point, $(X_{test}, Y_{test})$, where $X_{test}$ is the sample and $Y_{test}$ is its groundtruth label, CP enables coverage guarantees of the form~\cite{angelopoulos_cp_intro},
\begin{equation}
\label{eq1}
\mathbb{P}(Y_{test} \in \mathcal{C}(X_{test})) \ge 1 - \epsilon.
\end{equation}
According to~(\ref{eq1}),  $\epsilon = 0.05$ would mean that the probability of the correct class being within the predicted set is 95\%. 
The guarantee in~(\ref{eq1}) holds only if the training data, calibration data, and test data are exchangeable. Data points are said to be exchangeable when finite permutations of the data points do not change the joint probability distribution of the data~\cite{vovk_book}. Thus, the order in which the data is presented to the model does not reflect in the performance of the model. This allows establishing the guarantee in~(\ref{eq1}) through considering the error made by treating each data point as a new sample at the level~$\epsilon$~\cite{vovk_book}. It is interesting to note that exchangeability is a weaker assumption than identical and independently distributed (IID), which is more common in machine learning settings. IID random variables will be exchangeable because any permutation of the IID variables will yield the same joint distribution. However, exchangeable variables may or may not have independence though they will always be identically distributed.

A popular method for generating such conformal prediction sets is known as the adaptive prediction set (APS)~\cite{aps} method. In APS the training data is split into two, one for training the model and one for conformal calibration. The training process is subject to the network architecture not a subject of discussion of this article. During calibration, for each calibration data sample pair,~$(x_i, y_i)$, a conformity score, $\mathcal{S}_i = \mathcal{S}(x_i,y_i,u_i,\hat{f}^i)$ is calculated, where $\hat{f}^i = \hat{f}(x_i)$ is the vector of estimated probabilities, e.g., the softmax output of the trained model $f$. The estimated probability of an input $x_i$ belonging to class $c$ is denoted by $\hat{f}_c(x_i)$. The $u_i$ are  IID samples of $U \sim \text{Uniform}(0, 1)$. The score for each calibration data sample in the APS method is defined as 
\begin{equation}
\label{eq5}
\mathcal{S}(x,y,u,\hat{f}) = \text{min} \{\gamma \in[0, 1]: y \in \mathcal{C}(x,u,\hat{f},\gamma)\}.\end{equation} It is the smallest value of $\gamma \in [0,1]$  such that the prediction set generation function $\mathcal{C}$ contains the true class, $y$. The definition of $\mathcal{C}$ requires a quantile function $L$, defined as 
\begin{equation}
{\label{Eq3}}
    L(x, \hat{f}, \gamma) = \text{min} \{ c \in \{1,\dots, C \} : \sum_{c' = 1}^c \hat{f}_{(c')} \ge \gamma \}, 
\end{equation}
where $\hat{f}_{(\cdot)}$ is the descending order statistic for $\hat{f}$, i.e., $\hat{f}_{(1)}(x) \ge \hat{f}_{(2)}(x) \ge \cdots \ge  \hat{f}_{(C)}(x)$ and $C$ is the total number of classes. Therefore, for a given calibration sample the quantile function provides the number of classes to include in the prediction set till the threshold of $\gamma$ is reached. Finally the prediction set, $\mathcal{C}$ is generated as

\begin{equation}
\label{Eq4}
\begin{split}
&\mathcal{C}(x, u, \hat{f},\gamma) = \\
&\left\{ \begin{array}{ll}
   L-1  \text{ largest indices of } \hat{f}_c(x), &\text{if } u \le \Gamma(x,\hat{f}, \gamma) \\
L   \text{ largest indices of } \hat{f}_c(x), & \text{otherwise}, 
\end{array} \right.
\end{split}
\end{equation} where $u  \in [0, 1]$ and $\Gamma = \frac{1}{\hat{f}_L(x)}\left[ \sum_{c=1}^L \hat{f}_{(c)}(x) - \gamma\right].$ Upon calculation of the score for each calibration sample as in~(\ref{eq5}), $Q_{1-\epsilon}$ is set as the $\lceil (1- \epsilon)(1 + N) \rceil$ largest value of~$S = \{ \mathcal{S}_i\}_{i = 1}^N$, where $N$ is the number of calibration data samples. For the test sample $X_{test}$ the prediction set is then generated as \begin{equation}
    \label{eq6}
\mathcal{C}_{test} = \mathcal{C}(X_{test}, U_{test}, \hat{f}(X_{test}),Q_{1-\epsilon}),\end{equation} $U_{test} \sim \text{Uniform}(0, 1)$. The randomization due to $u$ in~(\ref{Eq4}) leads to the smallest prediction sets at the user desired coverage, $1 - \epsilon$~\cite{aps}. An example of conformal prediction with the APS method is provided in Fig.~\ref{cp_example}. The average cardinality of $\mathcal{C}_{test}$ in~(\ref{eq6}) is termed as the average prediction set size. A desirable property of the prediction set size is that it should be small for easy inputs and larger for difficult inputs. It reflects how uncertain the model is about its predictions. In the following section the average prediction set size will be used as an out-of-calibration detector for deep learning models.
\begin{figure}[!t]
\centering
\includegraphics[width = \columnwidth]{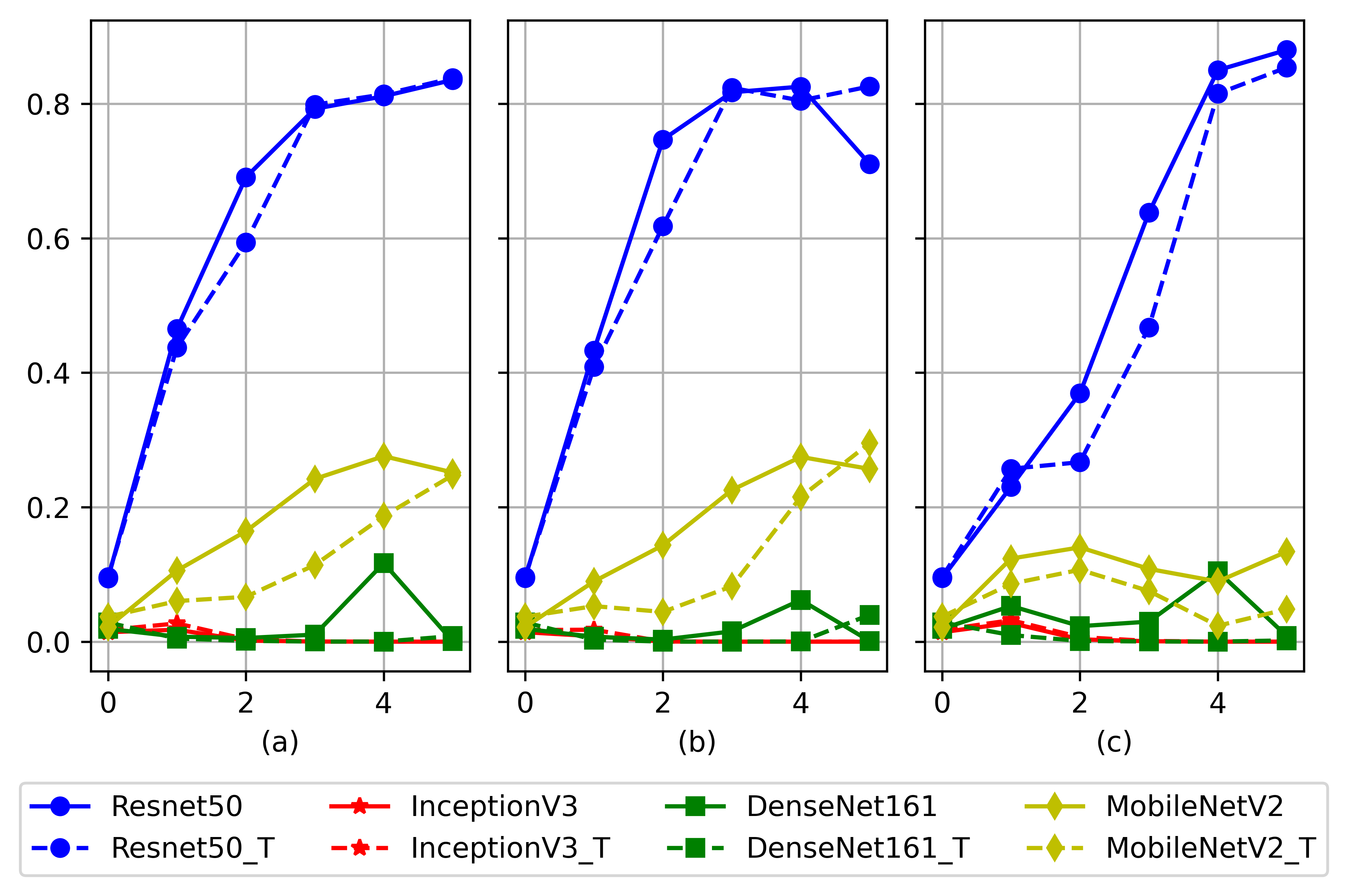}
\caption{Average normalized softmax entropy for different noise and models. (a) AWGN, (b) Shot noise, and (c) Impulse noise. The y-axis of each is the normalized softmax entropy and the x-axis is the noise severity as defined in~\cite{imagenetc}.}
\label{fig:entropy}
\end{figure}

\begin{figure}
\centering
\subfloat[]{\includegraphics[width=\columnwidth]{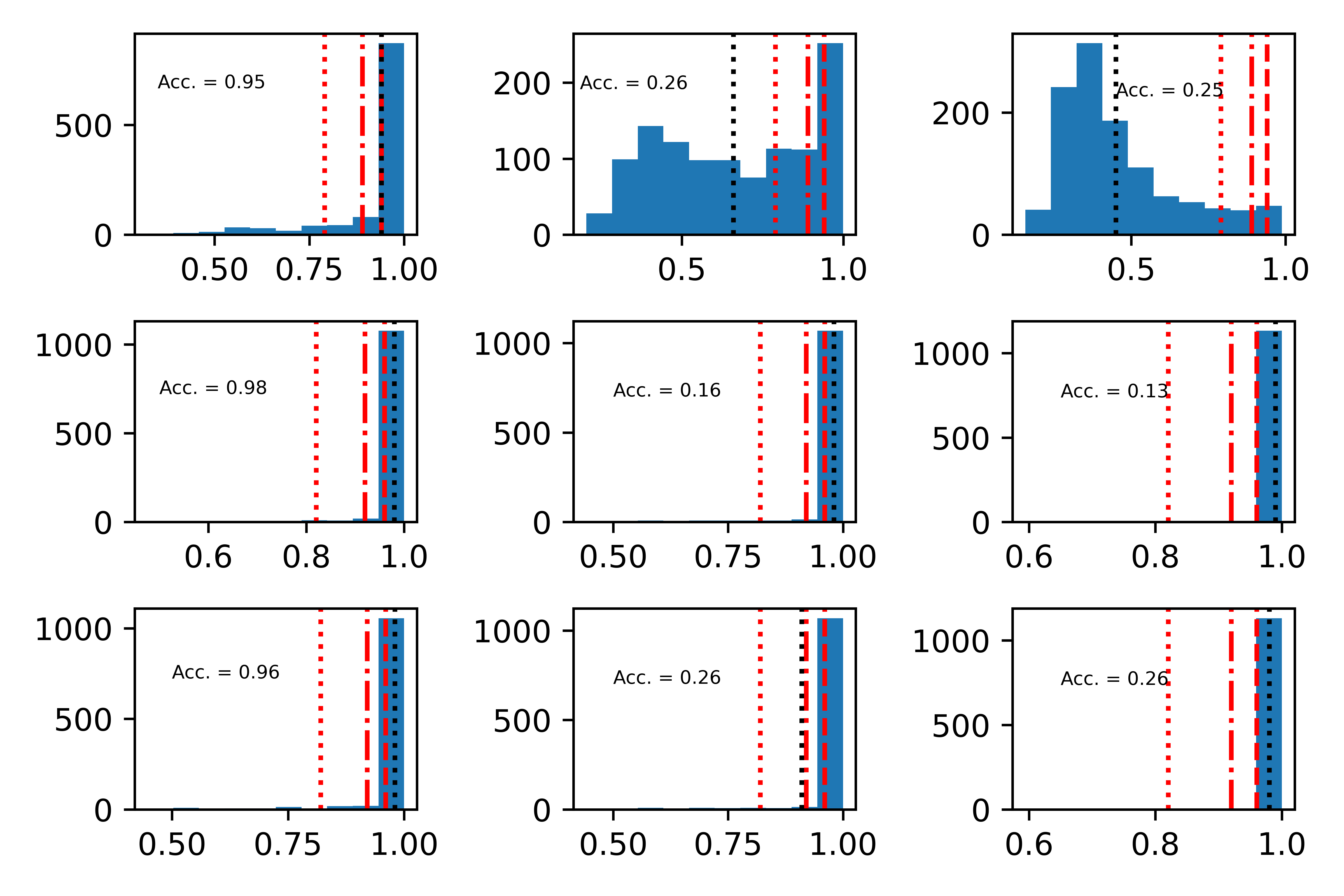}%
\label{fig:smx_awgn}}

\subfloat[]{\includegraphics[width=\columnwidth]{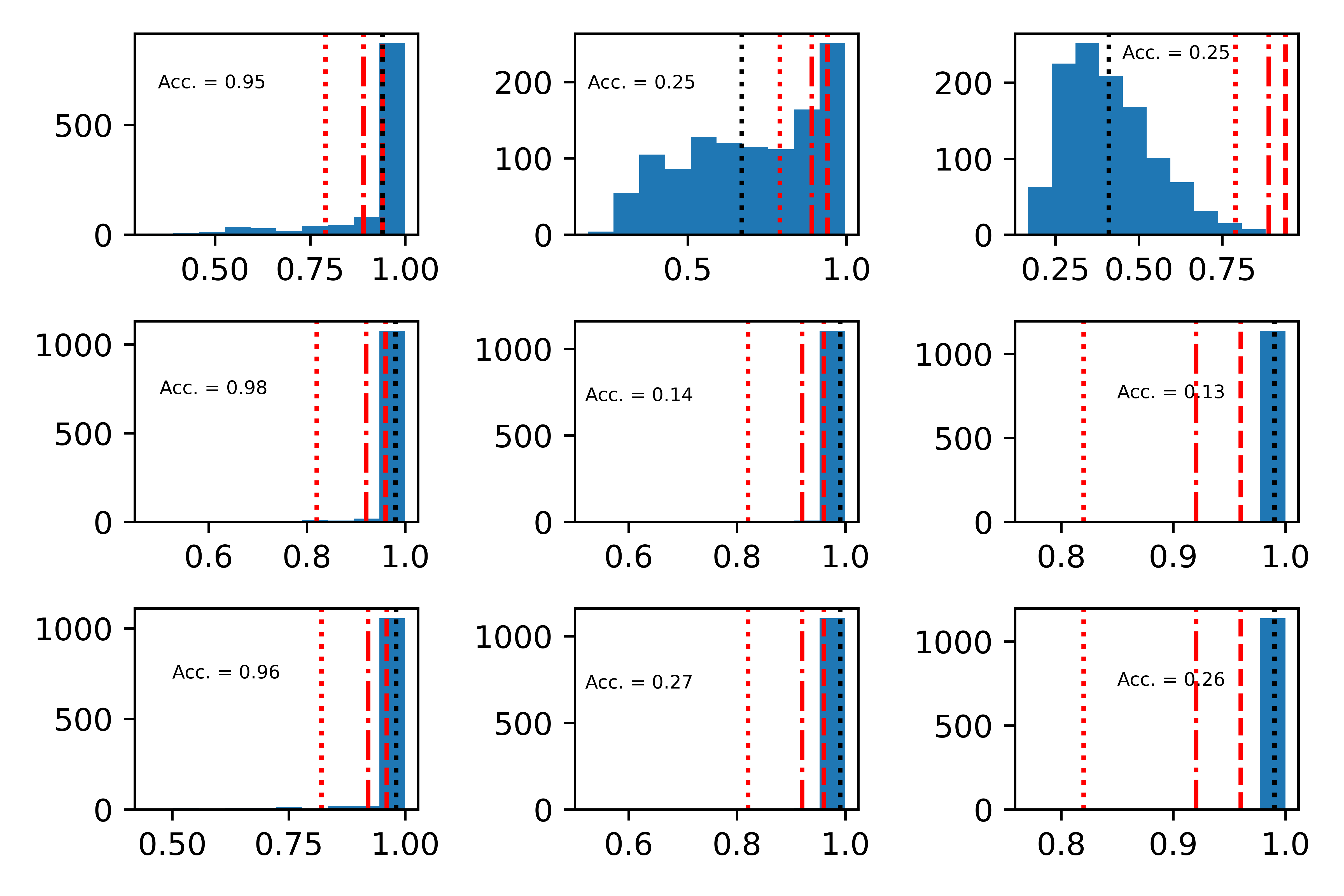}%
\label{fig:smx_shot}}

\subfloat[]{\includegraphics[width=\columnwidth]{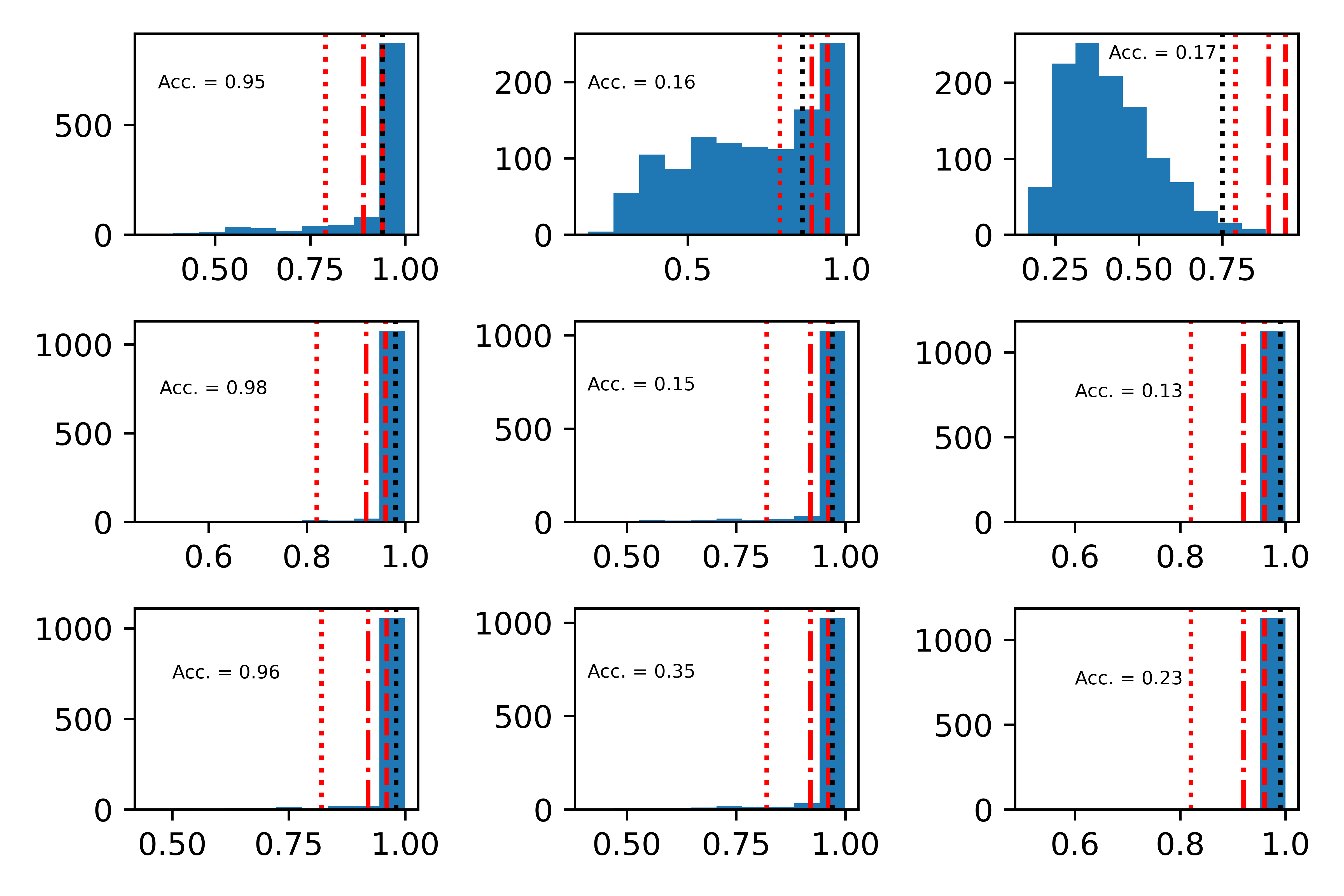}%
\label{fig:smx_impulse}}
\caption{Largest softmax histograms for different models and noise, (a) AWGN, (b) shot noise, and (c) impulse noise. In each subfigure \textit{Top Panel}: ResNet50, \textit{Middle Panel}: InceptionV3, and \textit{Bottom Panel}: MobileNetV2. 
\textit{Left Column}: clean validation data, \textit{Middle Column}: severity = 1, and \textit{Right Column}: severity = 2. The red \textit{dashed} line is $Q_{1- 0.05}$, \textit{dash-dot} is $Q_{1- 0.5}$, and \textit{dotted} is $Q_{1- 0.2}$ for APS. The black dotted line represents the average value of largest softmax.
}
\label{fig:softmaxHistogram}
\end{figure}

\section{Conformal prediction and uncertainty}
\label{sec:cp_n_uncertainty}
If exchangeability is violated, e.g., due to covariate shift~\cite{TibshiraniCovariateShift} or noise, then~(\ref{eq1}) does not hold. How do we detect, autonomously, that a covariate shift has taken place and the model is out-of-calibration? One method is to study the model uncertainty that can be quantified using the normalized softmax entropy (NSE) of a model, defined as  \begin{equation}
NSE = \frac{-\sum_{i=1}^C \hat{p}_ilog(\hat{p}_i)}{-log(\frac{1}{C})},
\label{eq:NSE}
\end{equation} where $\hat{p}_i$ is the estimated probability of the $i^{th}$ class. 
The higher the entropy the more uncertain the model. The average NSE 
for four popular models, Resnet50~\cite{eurosat2}, InceptionV3~\cite{inceptionv3}, Densenet161~\cite{densenet}, and MobileNetV2~\cite{mobilenetv2} and their corresponding temperature scaled version is shown in~Fig.~\ref{fig:entropy}. The NSE under different noise models~\cite{imagenetc}, specifically those common in remote sensing electronic systems~\cite{optSat_DC_book
}, such as additive white Gaussian noise (AWGN), shot, and impulse noise is also shown. Temperature scaling~\cite{tempscaling} is a method to calibrate models such that the estimated probabilities are better representation of the true probabilities. As is shown in Fig.~\ref{fig:entropy} the graphs for NSE for each model with and without temperature scaling follow each other very closely. For Resnet50 the NSE increases with increasing noise severity. Such a model can therefore be called an \emph{uncertain} model. On the other hand, the NSE values for InceptionV3 and Densenet161 remains constant near the low no noise value. Such models can said to be \emph{overconfident}. MobilenetV2 lies between these two extremes and shows a small increase in NSE with increasing noise at its input. 

Instead of entropy the APS algorithm, as in~(\ref{Eq3}), looks at the softmax values. An overconfident model tends to place importance on a single class irrespective of whether the prediction is correct or not. A distribution of the softmax outputs of such a network will peak near higher values. However, if a model is uncertain, then the model tries to output a flat softmax distribution that reflects the uncertainty in its predictions. An example is shown in Fig.~\ref{fig:softmaxHistogram}. The InceptionV3 and Densenet161 have very similar histograms and thus due to space limitations, Densenet161 histograms are not shown. With respect to the APS as described in Sec.~\ref{sec:cp}, the value of $Q_{1- \epsilon}$ for different~$\epsilon$ has been plotted in Fig.~\ref{fig:softmaxHistogram}. For clean data, the average softmax falls to the right of the corresponding $Q_{1-\epsilon}$. Thus, properly trained and conformally calibrated models generally have an average prediction set size close to 1. However, as noise increases, for the uncertain model, Resnet50, the average softmax moves to the left. Thus to reach the value of $\gamma = Q_{1-\epsilon}$ in~(\ref{Eq3}) more number of classes need to be included that increases the average prediction set size. This increase in the prediction set size can be used to indicate that the model is out-of-calibration. Once it is detected that the model input data is not exchangeable or there are deviations from the training and calibration data, it is clear that the network is experiencing an environment different from its training and calibration. On the other hand an overconfident model tends to put more probability on a single class, as a result the softmax histograms in the noisy case (middle and left columns of Fig~\ref{fig:softmaxHistogram}) look similar as in the clean case, the average softmax lies to the right of the $Q_{1-\epsilon}$ threshold to be used~(\ref{eq6}), and there is no significant change in the average prediction set size thus making it difficult to detect when the model is out-of-calibration. In the next section we study the change in prediction set size for the selected deep learning and noise models.

\section{Results}
\label{sec:Results}
EuroSAT~\cite{eurosat2} is a multispectral dataset curated from Sentinel-2 images for LULC that is used for simulations. It consists of 10 classes. For the purpose of this article the RGB version of the dataset was used. Out of the total 27000 samples, 21600 (80\%) were used for training the networks. The validation set consisted of 1140 samples and the calibration set had 3120 (11\%) samples. Common electronics noise encountered in remote sensing systems~\cite{optSat_DC_book} such as additive white Gaussian noise (AWGN), Poisson or shot noise, salt and pepper or impulse noise are considered. The noises are generated using methods in~\cite{imagenetc}. The APS implementation is used from the MAPIE~\cite{mapie} library. Upon acceptance for publication, the code will be released on Github. 

Table~\ref{Table:avg_pred_size_noise_all_models} summarizes the average prediction set size for the different models under different noise models and user defined tolerances, $\epsilon$. For the no noise cases, as $\epsilon$ increases the average prediction set size decreases. This can be also seen from Fig.~\ref{fig:softmaxHistogram}, the thresholds, $Q_{1 - \epsilon}$, lie to the left of the average softmax value (black dotted line). In some cases the average prediction set size is less than 1 indicating that the APS algorithm is generating null prediction sets. This can be understood from~(\ref{Eq4}). If the threshold is lower, e.g. if $\epsilon = 0.2$, a single class with a high softmax will lead to larger value of $\Gamma$, which leads to null sets with higher probability. However, the number of nulls do not vary much with the noise severity and thus is not used for analysis here. As noise is increased, the prediction set size for the uncertain Resnet50 model increases for all noises and severity. This can again be explained through Fig.~\ref{fig:softmaxHistogram}. With noise the average softmax value shifts to the left and for smaller value of $\epsilon$ a larger number of small valued softmax or more classes need to be considered to reach the calibration threshold leading to larger prediction sets. For shot noise the size decreases for the last two severity levels but they are significantly larger than the no noise case. Thus, an increase in the prediction set size can be used as an indicator for out-of-calibration detection. For InceptionV3 and Densenet161 the increase in the average prediction set size across $\epsilon$ and noise is negligible. Due to them being naturally overconfident in their estimates the average softmax is nearly equal in the noisy as well non-noisy case. It would be difficult to detect that the models are out-of-calibration using the prediction set size alone. MobileNetV2 lies somewhere in between the uncertain and overconfident models. For $\epsilon = 0.05 \ \text{and} \ 0.2$, MobileNetV2 behaves like the overconfident model, however for $\epsilon = 0.1$, one can observe that the average prediction set size steadily increases with anomalies similar to Resnet50. The increase is largest for AWGN ($\sim87\%$), the least for impulse noise ($\sim41\%$). Thus one can use such an analysis to develop a prediction set size threshold for their application to declare the corresponding deep learning model to be out-of-calibration.

\begin{table*}
\centering
\caption{ Average conformal prediction set size for different models under common noise types  \label{Table:avg_pred_size_noise_all_models}}

\begin{tabularx}{\textwidth} { 
  | >{\centering\arraybackslash}X 
  | >{\centering\arraybackslash}X 
  | >{\centering\arraybackslash}X 
  | >{\centering\arraybackslash}X
  | >{\centering\arraybackslash}X
  | >{\centering\arraybackslash}X
   | >{\centering\arraybackslash}X |}
\hline
\textbf{}& \textbf{Severity} & \textbf{Resnet50} & \textbf{InceptionV3} &  \textbf{Densenet161} &  \textbf{MobileNetV2}\\ \hline
&\multicolumn{5}{c|}{$\epsilon = 0.05$} \\ \hline
No noise & 0 & 1.246 & 1.005 & 1.024 & 1.005\\  \hline 
\multicolumn{1}{|c|}{\multirow{5}{*}{AWGN}} 
& 1 & 4.444 & 1.086 & 0.989 & 1.086 \\ \cline{2-6}
& 2 & 6.281  & 1.001 & 0.980 & 1.001  \\ \cline{2-6}
& 3 & 7.173 & 0.957 & 0.998 & 0.958 \\ \cline{2-6}
& 4 & 7.674  & 0.956 & 1.368 & 0.956  \\ \cline{2-6}
& 5 & 7.678 & 0.956 & 1.005 & 0.956  \\ \hhline {|=|=|=|=|=|=|} 
\multicolumn{1}{|c|}{\multirow{5}{*}{Shot Noise}} 
& 1 & 4.117 & 1.067 & 0.989 & 1.067 \\ \cline{2-6}
& 2 & 6.869  & 0.997 & 0.973 & 0.997  \\ \cline{2-6}
& 3 & 7.361 & 0.956 & 1.011 & 0.956 \\ \cline{2-6}
& 4 & 7.860  & 0.956 & 1.180 & 0.956  \\ \cline{2-6}
& 5 & 6.726 & 0.956 & 1.279 & 0.956  \\ \hhline {|=|=|=|=|=|=|} 
\multicolumn{1}{|c|}{\multirow{5}{*}{Impulse Noise}} 
& 1 & 2.394 & 1.052 & 1.132 & 1.056 \\ \cline{2-6}
& 2 & 3.681  & 0.967 & 1.038 &  0.967 \\ \cline{2-6}
& 3 & 5.982 & 0.957 &  1.063 & 0.957 \\ \cline{2-6}
& 4 & 7.768  & 0.956 &  1.321 & 0.956  \\ \cline{2-6}
& 5 & 8.035 & 0.956 & 1.010 & 0.956  \\ \hline

&\multicolumn{5}{c|}{$\epsilon = 0.1$} \\ \hline
No noise & - & 1.079 & 0.943 & 0.958 & 0.978\\  \hline 
\multicolumn{1}{|c|}{\multirow{5}{*}{AWGN}} 
& 1 & 3.221 & 1.056 & 0.938 & 1.212 \\ \cline{2-6}
& 2 & 4.828  & 1.024 & 0.931 & 1.404  \\ \cline{2-6}
& 3 & 5.708 & 0.925 & 0.928 & 1.706 \\ \cline{2-6}
& 4 & 6.319  & 0.915 & 0.928 & 1.832  \\ \cline{2-6}
& 5 & 6.393 & 0.914 & 0.939 & 1.739  \\ \hhline {|=|=|=|=|=|=|} 
\multicolumn{1}{|c|}{\multirow{5}{*}{Shot Noise}} 
& 1 & 2.890 & 0.985 & 0.933 & 1.173 \\ \cline{2-6}
& 2 & 5.370  & 1.033 & 0.928& 1.342  \\ \cline{2-6}
& 3 & 5.942 & 0.921 & 0.928 & 1.632 \\ \cline{2-6}
& 4 & 6.595  & 0.914 &0.928 & 1.809  \\ \cline{2-6}
& 5 & 5.143 & 0.914 & 1.022 & 1.740  \\ \hhline {|=|=|=|=|=|=|} 
\multicolumn{1}{|c|}{\multirow{5}{*}{Impulse Noise}} 
& 1 & 1.6551 & 0.979 & 0.947 & 1.328 \\ \cline{2-6}
& 2 & 2.635  & 0.922 & 0.931 & 1.383  \\ \cline{2-6}
& 3 & 4.446 & 0.915 &  0.928 & 1.257 \\ \cline{2-6}
& 4 & 6.461  & 0.915 &  0.928 & 1.131  \\ \cline{2-6}
& 5 & 6.927 & 0.914 & 0.932 & 1.246  \\ \hline

&\multicolumn{5}{c|}{$\epsilon = 0.2$} \\ \hline
No noise & - & 0.896 & 0.825 & 0.831 & 0.825\\  \hline 
\multicolumn{1}{|c|}{\multirow{5}{*}{AWGN}} 
& 1 & 2.086 & 0.907 & 0.814 & 0.907 \\ \cline{2-6}
& 2 & 3.255  &  0.849 & 0.813 & 0.849  \\ \cline{2-6}
& 3 & 4.023 & 0.807 &  0.818 & 0.807 \\ \cline{2-6}
& 4 &  4.502  & 0.806 &  1.001 & 0.806  \\ \cline{2-6}
& 5 & 4.582 & 0.806 &  0.844 & 0.806  \\ \hhline {|=|=|=|=|=|=|} 
\multicolumn{1}{|c|}{\multirow{5}{*}{Shot Noise}} 
& 1 & 1.842 & 0.898 & 0.813& 0.898 \\ \cline{2-6}
& 2 & 3.639  & 0.847 & 0.809& 0.847  \\ \cline{2-6}
& 3 & 4.247 & 0.806 & 0.826 & 0.806\\ \cline{2-6}
& 4 & 4.861  & 0.806 & 0.890 & 0.806  \\ \cline{2-6}
& 5 & 3.356 & 0.806 & 1.117 & 0.806  \\ \hhline {|=|=|=|=|=|=|} 
\multicolumn{1}{|c|}{\multirow{5}{*}{Impulse Noise}} 
& 1 & 1.108 & 0.847 & 0.883 & 0.847 \\ \cline{2-6}
& 2 & 1.688  & 0.811 & 0.836 & 0.811  \\ \cline{2-6}
& 3 & 2.842 & 0.808 &  0.845 & 0.808 \\ \cline{2-6}
& 4 & 4.775  & 0.806 &  0.969 & 0.806  \\ \cline{2-6}
& 5 & 5.270 & 0.806 & 0.836 & 0.806  \\ \hline
 
\end{tabularx}
\end{table*}

\section{Conclusion and future work}
\label{sec:conclusion}
The relation between model uncertainty and conformal prediction using the APS algorithm was studied in this work. Training models was not a focus of this paper, it is possible a better model exists for noisy or covariate shift scenarios, however this \emph{does not takeaway} the conclusion of this work that is in order to deploy neural networks as out-of-calibration detectors using conformal prediction, one must prefer an uncertain model instead of an overconfident one. Ability of distance aware Resnets for out-of-distribution (OOD) detection have been studied in~\cite{christophResnet}, which directly reflects their uncertainty studied in this article. Therefore, Resnets and to some extent MobileNets are good candidates to be used in onboard processing of data, as they can be adapted and integrated for health diagnostics/ monitoring pipelines of sensor systems~\cite{maik}. As CP algorithms~\cite{angelopoulos_cp_intro} utilize model logits or output softmax to design the prediction set generation procedure and a quantile based function to evaluate the class inclusion threshold $Q_{1 - \epsilon}$, the uncertainty concept can be extended to other conformal prediction algorithms as well. Here, only sensor noise was considered for model performance deterioration; in the future impact of intrinsic model parameters such as weight distortions will be studied for predictive uncertainty.

\bibliographystyle{IEEEtran}
\bibliography{IEEEabrv,main}

\end{document}